\documentclass{article}

\usepackage[final]{neurips_2025}
\usepackage{graphicx}
\usepackage{xcolor}
\usepackage[utf8]{inputenc}
\usepackage[T1]{fontenc}
\usepackage{algorithm}
\usepackage{algpseudocode}
\usepackage{amsmath}
\usepackage{enumitem}
\usepackage{listings}
\usepackage{xcolor}
\usepackage{url}
\usepackage{float}
\makeatletter
\renewcommand{\@notice}{}  % 禁用底部的 NeurIPS 标注框
\makeatother

\lstdefinelanguage{json}{
    basicstyle=\ttfamily\small,
    numbers=left,
    numberstyle=\tiny\color{gray},
    stepnumber=1,
    numbersep=5pt,
    showstringspaces=false,
    breaklines=true,
    frame=single,
    backgroundcolor=\color{gray!10},
    literate=
     *{0}{{{\color{blue}0}}}{1}
      {1}{{{\color{blue}1}}}{1}
      {2}{{{\color{blue}2}}}{1}
      {3}{{{\color{blue}3}}}{1}
      {4}{{{\color{blue}4}}}{1}
      {5}{{{\color{blue}5}}}{1}
      {6}{{{\color{blue}6}}}{1}
      {7}{{{\color{blue}7}}}{1}
      {8}{{{\color{blue}8}}}{1}
      {9}{{{\color{blue}9}}}{1}
      {:}{{{\color{black}:}}}{1}
      {,}{{{\color{black},}}}{1}
      {\{}{{{\color{black}\{}}}{1}
      {\}}{{{\color{black}\}}}}{1}
      {[}{{{\color{black}[}}}{1}
      {]}{{{\color{black}]}}}{1},
}
\usepackage[utf8]{inputenc} % allow utf-8 input
\usepackage[T1]{fontenc}    % use 8-bit T1 fonts
\usepackage{hyperref}       % hyperlinks
\usepackage{url}            % simple URL typesetting
\usepackage{booktabs}       % professional-quality tables
\usepackage{amsfonts}       % blackboard math symbols
\usepackage{nicefrac}       % compact symbols for 1/2, etc.
\usepackage{microtype}      % microtypography
\usepackage{xcolor}         % colors

\title{Structured Thinking Matters: Improving LLMs Generalization in Causal Inference Tasks}

\author{%
  Wentao Sun \\
  École Polytechnique \\
  \texttt{wentao.sun@polytechnique.edu} \\
  \texttt{wentao.sun@nokia.com} \\
  \And
  João Paulo Nogueira \\
  Institut Polytechnique de Paris \\
  \texttt{joaopaulo.fontoura@ip-paris.fr} \\
  \texttt{joao.fontoura\_nogueira@nokia.com} \\
  \And
  Alonso Silva \\
  Nokia Bell Labs \\
  \texttt{alonso.silva@nokia-bell-labs.com} \\
}

\begin{document}

\maketitle

\begin{abstract}
Despite remarkable advances in the field, LLMs remain unreliable in distinguishing causation from correlation~\citep{kiciman2024causal}. Recent results from the Corr2Cause dataset benchmark reveal that state-of-the-art LLMs—such as GPT-4 (F1 score: 29.08)—only marginally outperform random baselines (Random Uniform, F1 score: 20.38), indicating limited capacity of generalization~\citep{DBLP:conf/iclr/Jin0LPSMDS24}. To tackle this limitation, we propose a novel structured approach: rather than directly answering causal queries, we provide the model with the capability to structure its thinking by guiding the model to build a structured knowledge graph, systematically encoding the provided correlational premises, to answer the causal queries. This intermediate representation significantly enhances the model’s causal capabilities. Experiments on the test subset of the Corr2Cause dataset benchmark with Qwen3-32B model (reasoning model) show substantial gains over standard direct prompting methods, improving F1 scores from 32.71 to 48.26 (over 47.5\% relative increase), along with notable improvements in precision and recall. These results underscore the effectiveness of providing the model with the capability to structure its thinking and highlight its promising potential for broader generalization across diverse causal inference tasks.
\end{abstract}

\section{Introduction}

Inferring causation from correlation remains a fundamental yet challenging problem in science and artificial intelligence. While it is well-understood that correlation alone does not imply causation, reliably uncovering true causal relationships often requires structured reasoning or strong assumptions. Classical methods such as graphical models, the Peter–Clark (PC) algorithm~\citep{kalisch2007estimating}, and Greedy Equivalence Search (GES)~\citep{kalisch2007estimating,alonso2013scaling} offer principled frameworks for causal discovery from observational data. Human experts routinely construct causal diagrams to make reasoning transparent and systematic. However, whether LLMs~\citep{radford2019language,devlin2019bert} can perform causal inference with similar rigor remains an open question.

Recent studies have highlighted significant limitations in this regard. \citet{DBLP:conf/iclr/Jin0LPSMDS24} introduced the \textsc{Corr2Cause} benchmark to evaluate LLMs’ ability to infer causation from correlational input. Results were disappointing: even GPT-4 achieved only F1 score 29.08, while the best fine-tuned baseline (BART-MNLI) reached 33.38. Most models hovered near random or even below-chance performance. Extensive fine-tuning failed to yield robust generalization to novel causal structures, indicating a core deficiency in existing approaches.

We hypothesize that this underperformance stems from the lack of explicit structural reasoning in standard LLM workflows. Current models typically infer answers directly from textual patterns without systematically analyzing the underlying causal mechanisms—a process we term \textit{causal parroting}. In contrast, classical causal discovery and inference emphasizes constructing and reasoning over formal structures such as directed acyclic graphs (DAGs)~\citep{naser2025discovering}. Inspired by this, we propose a structured reasoning framework for LLMs that explicitly incorporates knowledge graph construction into the inference process.

Concretely, our method guides the LLM to first externalize its internal reasoning by generating an intermediate knowledge graph from correlational statements. This graph serves as a structured representation that constrains and guides the final causal judgment. Rather than guessing from surface cues, the model is prompted to identify paths, confounders, and potential interventions—just as a human analyst would.

To implement this framework, we leverage \textbf{Qwen3-32B}, an open-source LLM with native \textit{tool-calling capabilities}. Qwen3 can generate intermediate tool-call outputs and structured thinking traces, enabling the model to ``call'' a pseudo-tool that constructs a knowledge graph in a validated format. This mechanism forces structured thought before decision-making and ensures that reasoning remains transparent and verifiable. To our knowledge, Qwen3 is currently the only open-weight LLM that natively supports tool-calling for structured causal reasoning.\footnote{DeepSeek-R1 explicitly does not support tool-calling; see \url{https://github.com/deepseek-ai/DeepSeek-R1/issues/9}. Phi-4-Reasoning-Plus similarly lacks tool-call capabilities; see \url{https://huggingface.co/microsoft/Phi-4-reasoning-plus/discussions/13}.}

To evaluate the proposed method, we conducted experiments on the Corr2Cause benchmark for causal relation identification using the Qwen3-32B model (32 billion parameters). Zero-shot prompting achieved a precision of 31.61\%, recall of 33.89\%, F1 score of 32.71, and accuracy of 78.40\%. In contrast, our structured reasoning approach improves precision to 38.19\%, recall to 65.56\%, and F1 score to 48.26, while maintaining accuracy at 78.23\%. Compared to the baseline, this represents relative gains of 20.80\% in precision, 93.40\% in recall, and 47.50\% in F1 score, substantially reducing false positives. Furthermore, our approach demonstrates enhanced robustness on out-of-distribution queries, underscoring its strong generalizability.

\section{Related Work}
\label{gen_inst}

Causal inference in natural language processing (NLP) has attracted growing interest, but early efforts largely treated causality as a form of knowledge recall rather than reasoning~\citep{pearl2018book}. Many benchmarks focused on causal associations drawn from commonsense or factual data (e.g., cause-effect pairs in narratives or knowledge graphs), leading LLMs to behave as “causal parrots” that recite correlations seen in training data. Such approaches rely heavily on the coverage of causal pairs in the training corpus~\citep{DBLP:conf/iclr/Jin0LPSMDS24}, overlooking the need for formal reasoning to infer causation from mere correlation.

To push beyond knowledge-based inference,~\citet{DBLP:conf/iclr/Jin0LPSMDS24} introduced the Corr2Cause benchmark – the first large-scale dataset for pure causal inference in NLP. Corr2Cause poses a rigorous challenge: given a set of correlational statements (e.g., “A is correlated with B; C is independent of A, etc.”) and a hypothesized causal relation, the model must decide if the causation claim is valid. The dataset contains over 200,000 examples generated using formal causal discovery rules, ensuring that each instance tests whether an LLM can deduce the underlying knowledge graph consistent with observed correlations. This benchmark shifts the focus from recalling commonsense causal facts to reasoning about causality from statistical evidence.

Performance on Corr2Cause has revealed a significant gap in current models’ abilities. \citet{DBLP:conf/iclr/Jin0LPSMDS24} evaluated 17 different LLMs (including BERT-family models, GPT-3.5, GPT-4, and instruction-tuned variants) on this benchmark, and none achieved satisfactory results. In fact, across these models the F1 scores hover near the random-guess baseline~\citep{DBLP:conf/iclr/Jin0LPSMDS24}. Even GPT-4 model performs only marginally above chance, with many queries answered incorrectly. Some models fare worse than random, underscoring the fundamental difficulty of the task. For example, off-the-shelf natural language inference models like BERT and RoBERTa (fine-tuned on MNLI) often default to predicting “no causation” for every case, yielding extremely low recall of true causal relations (F1 well under 5\% in such cases). These results make clear that simply scaling up model size or using generic training is insufficient for causal inference – existing LLMs overwhelmingly struggle to infer causation from correlation.

One obvious remedy is to fine-tune LLMs on the Corr2Cause training data. Finetuning does help to an extent: for instance, a RoBERTa-Large model finetuned on Corr2Cause can reach an impressive ~95\% F1 on the original test set. However, this success is deceptive – the model fails to generalize beyond the narrow patterns it learned. When the correlational statements are paraphrased or variable names changed (i.e. out-of-distribution scenarios), performance plummets dramatically.~\citet{DBLP:conf/iclr/Jin0LPSMDS24} report that even the best finetuned model loses much of its accuracy under such perturbations, indicating a brittle, overfit reasoning capability. In short, in-distribution training can inflate performance, but it does not solve the core reasoning problem: the model still lacks a robust causal inference strategy and instead has likely memorized superficial cues.

The difficulty of Corr2Cause has spurred follow-up research seeking more principled solutions. One notable line of work is to incorporate explicit causal reasoning steps into the LLM’s prompting procedure. For example,~\citet{sgouritsa2024prompting} propose a multi-step prompting strategy (PC-SubQ) that breaks down the Corr2Cause problem into a sequence of sub-questions aligned with the classical PC causal discovery algorithm. By guiding the LLM through each step of a formal algorithm (e.g., identifying independencies, colliders, etc.), their approach significantly improves performance across several LLMs and makes the results more robust to variations in phrasing. This indicates that strategic prompting – forcing the model to emulate a step-by-step causal discovery process – can yield better results than end-to-end black-box reasoning.

Another promising direction is to enhance LLMs through external tool usage. Tool-augmented frameworks such as TALM~\citep{parisi2022talm} allow LLMs to dynamically call external APIs and incorporate the outputs into their generation process. TALM introduces a self-play mechanism to bootstrap tool usage from few demonstrations, and has shown effectiveness on both knowledge-intensive and reasoning tasks. This idea of augmenting LLMs with explicit tool invocation aligns closely with our method, which equips the model with the ability to invoke a specialized graph-construction tool for building structured causal representations. By transforming the causal inference task into a structured tool calls, we aim to bridge the gap between symbolic reasoning and language understanding. We detail our approach in the following section.

\section{Methodology}
\label{sec:method}

In order to enable reliable causal reasoning in LLMs, we design a pipeline that explicitly separates
\emph{causal structure induction} from \emph{causal conclusion drawing}. First, given the correlational
premises, the model is prompted to construct a structured graph that captures a plausible underlying
causal model. Second, the model performs structure-aware causal inference by consulting this graph
to answer the causal query. By interposing a graph representation, our method forces the LLM to
articulate the assumed causal relationships before finalizing its answer, thereby encouraging more
systematic reasoning.

\subsection{Knowledge Graph Generation}
\label{sec:graph_generation}

In the first stage, the model constructs a \textbf{skeleton graph} that captures the observed statistical dependencies. We prompt the model with a list of observed correlations and independencies, and instruct it to output a JSON representation listing the variables (nodes) and the dependency edges:

\begin{lstlisting}[language=json]
{
  "nodes": ["X", "Y", "Z"],
  "edges": [
    {"source": "X", "target": "Y", "label": "correlates with"},
    {"source": "X", "target": "Z", "label": "independent of"}
  ]
}
\end{lstlisting}

For example, if the premise states “X is correlates with Y, and Z is independent of X,” the output graph contains an undirected edge between X and Y, and no edge between X and Z. At this stage, causal directions remain unspecified; the graph merely reflects the statistical dependencies implied by the premise.

\begin{figure}[H]
  \centering
  % 裁掉图片左右各 50pt 的空白，并把宽度调成文字区的 0.9
  \includegraphics[
    width=0.9\linewidth,
    trim=80 0 0 0,
    clip
  ]{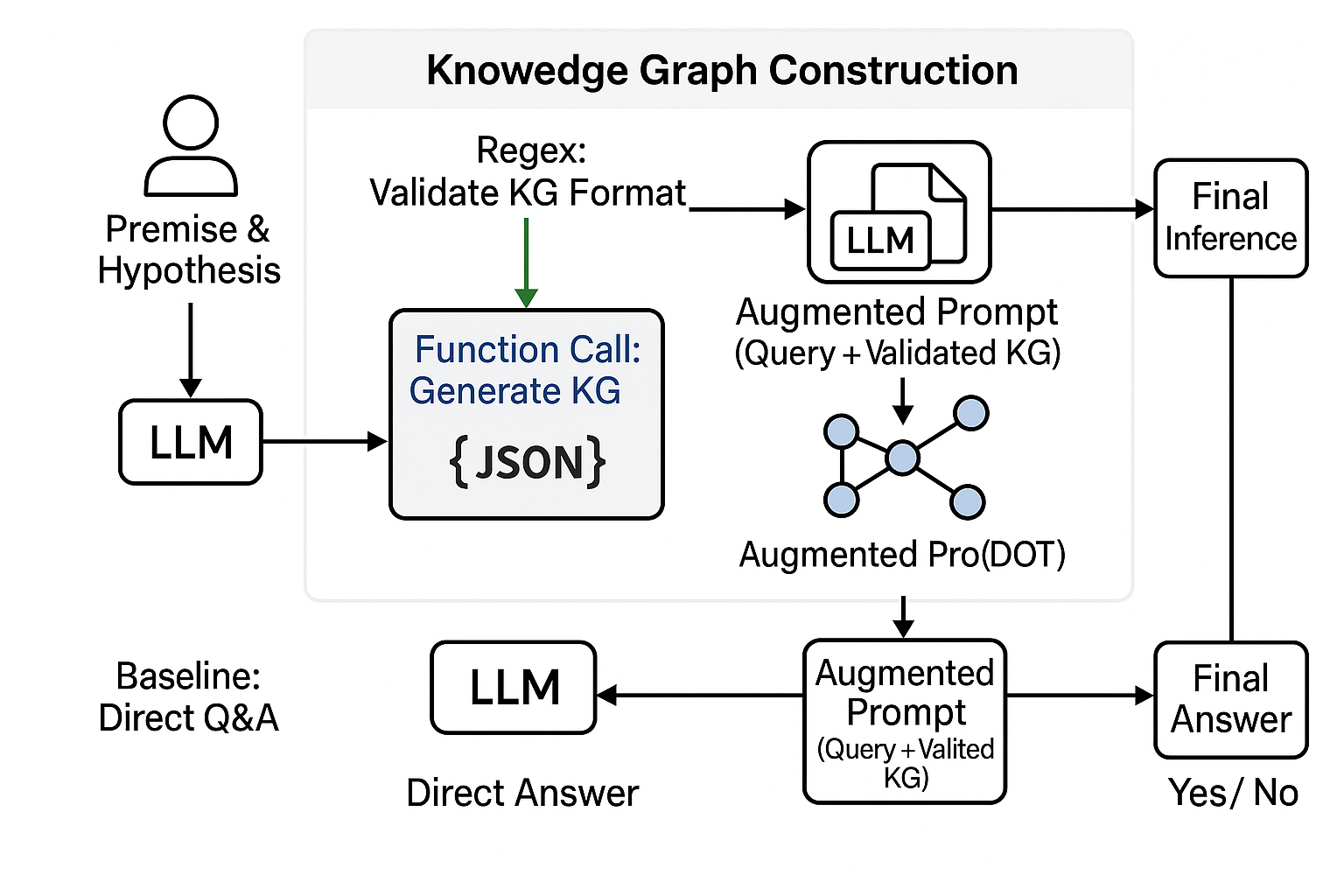}
  \caption{Overview of our structured causal reasoning pipeline.}
  \label{fig:framework}
\end{figure}

To ensure the output graph is well-structured, we employ an OpenAI-style tool calling approach
and regex-based output control. The model is given a predefined JSON schema (for instance, a synthetic
function signature requiring a list of nodes and edges), which guides it to produce a JSON answer
conforming to that schema. This structured outputs technique prevents the LLM from drifting into
free-form text, effectively sandboxing its response as a machine-readable graph.

By parsing the correlational input into a graph, causal directions are explicitly defined: the model lays out hypotheses of which variables act as causes and which act as effects. Importantly, this step leverages the LLM’s capability
to perform implicit causal discovery through natural language understanding.
\subsection{Structured Output via Tool Calling}
\label{sec:tool_call}

To guarantee that every induced graph is exactly valid JSON \emph{and} conforms to our schema, we enforce three complementary components:

\begin{itemize}[leftmargin=*, itemsep=0.6em]
  \item \textbf{Schema definition.}  
    We capture the skeleton graph with a strongly‐typed Pydantic model:
    \[
      \texttt{KnowledgeGraph(nodes: List[Node], edges: List[Edge])},
    \]
    where each \texttt{Node} has an integer ID and a label, and each \texttt{Edge} has source/target IDs plus a descriptive label (e.g.\ “\texttt{correlates with}” or “\texttt{independent given A and B}”).

  \item \textbf{Tool conversion.}  
    The Pydantic schema is converted into an OpenAI function (“tool”) via
    \[
      \texttt{convert\_to\_openai\_tool(KnowledgeGraph)},
    \]
    automatically generating the precise JSON signature that the LLM must adhere to when emitting its graph.

  \item \textbf{Output enforcement.}  
    We compile the JSON schema into a whitespace‐insensitive regular expression and embed it in our control tags:
    \begin{itemize}[leftmargin=1.5em]
      \item Serialize the Pydantic schema (\texttt{convert\_json\_schema\_to\_str}).
      \item Build a regex from it (\texttt{build\_regex\_from\_schema}).
      \item Enclose the pattern in \verb|<think>…</think>| and \verb|<tool_call>…</tool_call>|.
      \item Inject the resulting regex into the model’s logits processor (\texttt{RegexLogitsProcessor}).
    \end{itemize}
    This pipeline prevents any free‐form drift: every model output is guaranteed to match our schema exactly.
\end{itemize}
\subsection{Structure-Aware Causal Inference}
\label{sec:inference}

In the second stage, the LLM uses the generated graph to inform its final answer to the causal
question or hypothesis. We provide the model with the previously constructed graph (for example,
by including the JSON graph in the prompt or via a tool interface) alongside the original causal
query. The model’s task is now framed as: given this candidate knowledge graph, determine whether the
stated hypothesis holds. Because the model has an explicit graph to consult, its reasoning can be
grounded in the graph’s structure rather than relying on superficial cues from the textual premise.

Concretely, the model checks the hypothesis against the graph. For instance, if the hypothesis is
“X directly causes Z,” the model will verify whether there is a direct path from X to Z in the graph and no confounding paths that violate the claim). The structured graph provides a clear criterion:
the hypothesis is true if and only if it is consistent with the graph’s causal relationships. The LLM,
guided by the graph, can thus make a determination with greater confidence.

In our prompt design, we may ask the model to explain its answer using the graph – for example,
“According to the graph, $X$ causes $Y$ which in turn causes $Z$, so $X$ indirectly affects $Z$ (not
a direct cause),” thereby ensuring the model’s reasoning remains faithful to the structure.

By making the model structure-aware, this step markedly improves reliability. The LLM is no longer
attempting to implicitly juggle correlation and causation in a single pass; instead, it has a concrete
graph to reference. This separation of concerns reduces reasoning complexity: the first stage distills
the raw text into an abstract causal form, and the second stage is a straightforward evaluation of the
hypothesis against that form.

Our approach thus injects a form of algorithmic transparency into the LLM’s reasoning process,
as the intermediate graph can be inspected and even evaluated by humans or other systems. In
summary, the methodology ensures that any causal conclusion the model draws is backed by an
explicit structural justification, mitigating the risk of spurious correlations leading the model astray.

\section{Experimental Setup}
\label{others}

We conduct our experiments on the Corr2Cause benchmark, a recent dataset designed to test the pure causal‐inference capabilities of LLMs. Each instance includes a set of correlational statements (the \textit{premise}) and a causal hypothesis. The task is to determine whether the hypothesis causally follows from the given premise. The dataset is heavily imbalanced, containing approximately 80\% negative examples (``No'') and 20\% positive examples (``Yes''). Due to this skew, accuracy becomes a less informative metric—a trivial classifier that always predicts ``No'' would achieve approximately 80\% accuracy. Therefore, we emphasize the F$_1$ score, particularly for the minority ``Yes'' class, to more fairly assess model performance under class imbalance.

As a baseline, we employ a direct zero‐shot prompting approach similar to Jin et al. The model is given a natural‐language question of the form:
\begin{verbatim}
Question: [Premise]. 
Can we deduce the following: [Hypothesis]? 
Just answer "Yes" or "No".
\end{verbatim}
The model then responds with either ``Yes'' or ``No''. This setup evaluates the model's raw ability to distinguish causation from correlation without any structured reasoning or intermediate representations.

Our proposed method enhances reasoning through a structured tool‐calling framework. Instead of directly answering the yes/no question, the model is first prompted to generate a structured knowledge graph representing the relationships between variables described in the premise. This graph is constructed using a predefined tool‐calling schema, capturing correlations, independencies, and other potential links. Subsequently, the model reasons over this structured graph to assess whether the hypothesized causal relation follows. This two‐step process—(1) parsing the premise into a formal graph and (2) applying reasoning over the graph—this can be seen as a kind of \textit{chain‐of‐thought}, improving transparency and separation of comprehension and inference.

\begin{algorithm}
\caption{Causal Inference with KG‐Augmented LLM}
\begin{algorithmic}[1]
\Require Dataset $\mathcal{D} = \{(x_i, y_i)\}$, where $x_i =$ “Premise + Hypothesis”
\Ensure Predicted labels $\hat{y}_i \in \{\text{Yes}, \text{No}\}$
\State Load pretrained LLM with tool‐calling capability
\State Convert KG schema into Regex‐constrained JSON pattern
\ForAll{example $x_i$ in filtered subset of $\mathcal{D}$}
    \State Split $x_i$ into \textsf{Premise} and \textsf{Hypothesis}
    \State Build prompt $p_i$ to instruct LLM to generate KG before answering
    \State Use Regex‐constrained decoding to use tool-calling to generate valid KG JSON from $p_i$
    \State Parse JSON output $\text{KG}_i$ as $(\text{nodes}, \text{edges})$
    \State Convert $\text{KG}_i$ into DOT format string $\text{DOT}_i$
    \State Give the result of the tool call to the LLM: $q_i = p_i + \text{KG}_i + \text{DOT}_i$
    \State Query LLM with $q_i$ to obtain final prediction $\hat{y}_i$
\EndFor
\State \Return All $\hat{y}_i$
\end{algorithmic}
\end{algorithm}

We evaluate two LLMs from the Qwen family, both with 32 billion parameters. Qwen2.5‐32B is a previous‐generation model without specialized reasoning capabilities, while Qwen3-32B is a newer model with improved structured reasoning abilities. Both models are evaluated in a zero‐shot setting, with no fine‐tuning on Corr2Cause. Qwen3 is designed to handle tool‐use and chain‐of‐thought prompting more effectively, enabling it to follow our structured reasoning pipeline~\citep{qwen3_technical_report}. In contrast, Qwen2.5 is included as a reference model and is expected to struggle with such complex multi‐step prompts\citep{yang2024qwen2}.

To assess the impact of KG serialization syntax on LLM comprehension, we introduce an auxiliary experiment using a smaller variant, Qwen3-4B, while holding the rest of the pipeline constant. We generate KGs in three different edge‐notation styles:

\begin{enumerate}[label=\textbf{Style \arabic*:}]
  \item \texttt{a -> b [dir=none]}  
  \item \texttt{a -> b + b -> a}  
  \item \texttt{a -- b}  
\end{enumerate}

Style 1 uses a directed arrow with an explicit `dir=none' attribute to signal an undirected relation; style 2 encodes an undirected link via two opposing directed edges; and style 3 employs a simple undirected edge notation. By comparing F$_1$, precision, and recall on the positive class across these styles, we identify which syntax the model parses most reliably. Results in Table~\ref{tab:style-eval} show that Style 2 yields the best F$_1$ for the ``Yes'' class, and we adopt this notation in our main KG generation.

We report F$_1$ score (main metric), recall, precision, and accuracy on the Corr2Cause test set. Our key interest lies in the model’s ability to correctly identify positive (``Yes'') cases of true causation, without being overwhelmed by the dominant ``No'' class. All evaluations are conducted in the zero‐shot setting: the models are presented with each instance for the first time and must rely entirely on their pre‐trained knowledge and reasoning capabilities.

\section{Results and Analysis}

\subsection{Graph Style Evaluation (Qwen3-4B)}

Before analyzing the full performance of different models and prompting strategies, we first investigate how the serialization format of the generated knowledge graph affects model comprehension. We conduct a controlled experiment using a smaller model variant (Qwen3-4B), holding all other components fixed while varying the KG output style.

Table~\ref{tab:style-eval} reports the results of three KG notations: 

Among these, \texttt{Style 2} consistently yields the highest F$_1$ score (0.3617) and recall (0.5075), suggesting that the model parses this format more reliably. We therefore adopt this style across all subsequent experiments and generate visual KGs using Graphviz via the code call \texttt{dot.edge(src, tgt, label=lbl)}.

\begin{table}[H]
\centering
\caption{Performance of Qwen3-4B under different KG styles.}
\label{tab:style-eval}
\begin{tabular}{lcccc}
\toprule
\textbf{Style} & \textbf{Accuracy} & \textbf{Precision} & \textbf{Recall} & \textbf{F1} \\
\midrule
Style 1 (\texttt{a -> b [dir=none]}) & 0.7447 & 0.2632 & 0.4478 & 0.3315 \\
Style 2 (\texttt{a -> b + b -> a})   & \textbf{0.7468} & \textbf{0.2810} & \textbf{0.5075} & \textbf{0.3617} \\
Style 3 (\texttt{a -- b})            & 0.7321 & 0.2500 & 0.4478 & 0.3209 \\
\bottomrule
\end{tabular}
\end{table}

\subsection{Main Performance Comparison}

We now turn to the evaluation of our full structured reasoning framework on the Corr2Cause benchmark.

Before comparing different prompting strategies on Qwen variants, we benchmark our best‐performing model against two strong off‐the‐shelf baselines: GPT‐4 and BART MNLI. Table~\ref{tab:baseline-comparison} summarizes their performance alongside Qwen3-32B. Both GPT‐4 and BART MNLI achieve only moderate F$_1$ scores (29.08 and 33.38), with GPT‐4 suffering from low precision and BART MNLI trading precision for recall. In contrast, Qwen3-32B under our structured framework achieves a much higher F$_1$ of 48.26, driven by strong recall (65.56) and improved precision (38.19), while maintaining competitive accuracy.

\begin{table}[H]
\centering
\caption{Performance of off‐the‐shelf baselines vs.\ Qwen3-32B (Structured Reasoning) on Corr2Cause.}
\label{tab:baseline-comparison}
\begin{tabular}{l c c c c}
\toprule
\textbf{Model} & \textbf{F1} & \textbf{Precision} & \textbf{Recall} & \textbf{Accuracy} \\
\midrule
GPT-4                      & 29.08 & 20.92 & 47.66 & 64.60 \\
BART MNLI                  & 33.38 & 31.59 & 35.38 & \textbf{78.50} \\
Qwen3-32B (Structured)     & \textbf{48.26} & \textbf{38.19} & \textbf{65.56} & 78.23 \\
\bottomrule
\end{tabular}
\end{table}

We further compare Qwen2.5 and Qwen3 under direct vs.\ structured approach (Table~\ref{tab:main-results}). Structured reasoning notably boosts Qwen3’s F$_1$ score from 32.71 to 48.26, nearly doubling recall and improving precision. Qwen2.5 also benefits modestly, but its lack of structural understanding limits gains.

\begin{table}[H]
\centering
\caption{Evaluation results on Corr2Cause using two prompting strategies: direct vs.\ structured reasoning.}
\label{tab:main-results}
\begin{tabular}{l l c c c c}
\toprule
\textbf{Model (32B)} & \textbf{Approach}              & \textbf{F1}   & \textbf{Recall} & \textbf{Precision} & \textbf{Acc} \\
\midrule
Qwen2.5-32B & Unstructured (Baseline)      & 19.51        & 17.27          & 22.43             & \textbf{81.60} \\
Qwen2.5-32B & Structured                   & \textbf{24.71} & \textbf{23.33} & \textbf{26.25}    & 77.97          \\
\midrule
Qwen3-32B   & Unstructured (Baseline)      & 32.71        & 33.89          & 31.61             & \textbf{78.40} \\
Qwen3-32B   & Structured                   & \textbf{48.26} & \textbf{65.56} & \textbf{38.19}    & 78.23          \\
\bottomrule
\end{tabular}
\end{table}

Overall, Qwen3 with intermediate KG reasoning achieves superior precision and recall, suggesting it better distinguishes genuine causal relationships. Its use of DAG structures during inference leads to clearer and more interpretable answers, often avoiding logical contradictions seen in baseline responses.

\section{Conclusions and Discussion} We have presented a novel structured reasoning framework for causal inference that guides LLMs to first construct explicit knowledge-graph representations of correlational premises before making causal judgments. Our approach leverages the Qwen2.5-32B model and the recent Qwen3-32B model equipped with tool-calling, allowing the LLM to output intermediate, machine-validated knowledge graphs. This structured pipeline transforms the Corr2Cause task into a graph reasoning problem, effectively decoupling correlation analysis from causal judgment.

Empirically, this process significantly improves performance: on the Corr2Cause benchmark, our method achieves substantially higher F1 and recall scores than baseline models, and it is markedly more robust to out-of-distribution query perturbations. Compared to conventional end-to-end prompting~\citep{xu2023making} or chain-of-thought~\citep{wei2022chain} strategies, the graph-based pipeline has several key advantages. The intermediate graph representation makes the model's reasoning more transparent and allows explicit consistency checks, reducing reliance on spurious linguistic patterns in the input. Grounding the model's decisions in this explicit structure leads to more reliable outputs and higher generalization, especially in challenging, distribution-shifted settings. 

We observe that the graph-guided method maintains its accuracy and recall even when variable names or textual descriptions are paraphrased or replaced, demonstrating strong robustness to adversarial query variations. 

Despite these benefits, our approach has limitations. Generating detailed knowledge graphs for each query is computationally intensive and requires careful design of the graph schema. Errors in graph generation or tool invocation can cascade into incorrect causal judgments. The framework's effectiveness depends on the fidelity of the knowledge schema and the model's ability to correctly populate it, which may degrade on very complex or ambiguous inputs. We also observe that adversarial or unusually phrased inputs can sometimes confound the graph construction process, indicating potential brittleness in those edge cases. 

Looking forward, there are several promising directions for future work. One avenue is to develop automated verification or validation mechanisms for the generated graphs, for example by integrating symbolic or logical consistency checks to catch erroneous edges or missing relations. Extending the structured reasoning approach to other complex reasoning tasks (such as multi-hop question answering, scientific reasoning, or richer causal discovery scenarios) could further demonstrate its generality. Integrating the LLM-based graph reasoning with external symbolic inference engines or knowledge bases is another promising direction, which may improve factual grounding and inference efficiency. Additionally, optimizing the tool-calling pipeline for efficiency (e.g., by streamlining the graph construction or pruning irrelevant details) or exploring joint training of graph schemas and generation could address current scalability and robustness challenges. 

In summary, our work underscores the importance of structured, modular reasoning in improving LLM performance on causally challenging tasks. By explicitly bridging correlational input and causal inference through an interpretable graph-based intermediate, we enable better generalization and robustness. Our results confirm that encouraging structured thinking in LLMs significantly enhances their reasoning capabilities. We hope this study inspires further exploration into hybrid neuro-symbolic approaches that leverage the strengths of LLMs while ensuring more faithful and verifiable reasoning~\citep{amizadeh2020neuro}.

\newpage

\small
\bibliographystyle{plainnat}  
\bibliography{reference}

\end{document}